\documentclass[10pt,twocolumn,letterpaper]{article}

\usepackage{cvpr_arxiv}
\usepackage{times}
\usepackage{epsfig}
\usepackage{graphicx}
\usepackage{amsmath}
\usepackage{amssymb}
\usepackage[]{algorithm2e}
\usepackage{subfigure}
\usepackage{placeins}

\usepackage[pagebackref=true,breaklinks=true,colorlinks,bookmarks=false]{hyperref}
\newcommand\inlineeqno{\stepcounter{equation}\ (\theequation)}

\cvprfinalcopy % *** Uncomment this line for the final submission

\def\BibPath{.}  

\def\cvprPaperID{1} % *** Enter the CVPR Paper ID here

\begin{document}

%%%%%%%%% TITLE
\title{UAV-based Autonomous Image Acquisition\\with Multi-View Stereo Quality Assurance by Confidence Prediction}

\author{
Christian Mostegel
~~~~~~~~~~
Markus Rumpler
~~~~~~~~~~
Friedrich Fraundorfer
~~~~~~~~~~
Horst Bischof\\
Institute for Computer Graphics and Vision, Graz University of Technology\thanks{The research leading to these results has received funding from the EC FP7 project 3D-PITOTI (ICT-2011-600545)
and from the Austrian Research Promotion Agency (FFG) as Vision+ project 836630 and together with OMICRON electronics GmbH as Bridge1 project 843450.}\\
{\tt\small \{surname\}@icg.tugraz.at}
}
\maketitle
%\thispagestyle{empty}

%%%%%%%%% ABSTRACT
\begin{abstract}
In this paper we present an autonomous system for acquiring close-range high-resolution images that 
maximize the quality of a later-on 3D reconstruction with respect
to coverage, ground resolution and 3D uncertainty.
In contrast to previous work, 
our system uses the already acquired images to predict the confidence 
in the output of a dense multi-view stereo approach without executing it.
This confidence encodes the likelihood of a successful reconstruction
with respect to the observed scene and potential camera constellations.
Our prediction module runs in real-time and 
can be trained without any externally recorded ground truth.
We use the confidence prediction 
for on-site quality assurance and for planning further views that
are tailored for a specific multi-view stereo approach
with respect to the given scene.
We demonstrate the capabilities of our approach with an autonomous Unmanned Aerial Vehicle (UAV) 
in a challenging outdoor scenario.
\end{abstract}
\vspace{-12pt}
\section{Introduction}

In this paper, we address the problem of UAV-based image acquisition for dense monocular 3D reconstruction
with high-resolution images at close range.
The aim is to acquire images in such a way that 
they are suited for processing with an offline dense multi-view stereo (MVS) algorithm,
 while at the same time fulfilling a set of quality requirements.
 These requirements include coverage, ground resolution and 3D accuracy and
 can be assessed geometrically.
 However, determining how well the images are suited 
for a specific MVS algorithm is much harder to model.
To extract depth from 2D images, MVS approaches have to
establish correspondences between the images.
To solve this challenging task every MVS approach has to
make some assumptions.
These assumptions vary from approach to approach, but
the most popular assumptions include saliency, local planarity and a static environment.
If some of these assumptions are violated the MVS algorithm will not be able to reconstruct the scene correctly.
Up to now,
this problem was widely ignored by monocular image acquisition approaches~\cite{dunn09nbv,hoppe12nbv,schmid12view_planning_mav,munkelt10mvp,hossein13,martin16,pistellato15},
which often leads to missing parts in the resulting 3D reconstructions~\cite{schmid12view_planning_mav,hoppe12nbv}.
In this work, we propose a solution for this problem via machine learning.
The main idea is to predict 
how well the acquired images are suited for the dense MVS algorithm directly during the acquisition.
While this is already useful for quality assurance, we take this idea one step further
and use the acquired images to plan the optimal camera constellation with respect to the observed scene structure.
Within this context, we demonstrate that the likelihood of a successful 3D reconstruction
depends on the combination of scene structure,
triangulation angle and the used MVS algorithm.
We further refer to the prediction of this likelihood as \emph{MVS confidence prediction}.
\begin{figure}[top]
   \vspace{-10pt}
  %\centering
 % \hspace{-10pt}
\includegraphics[width=1\columnwidth]{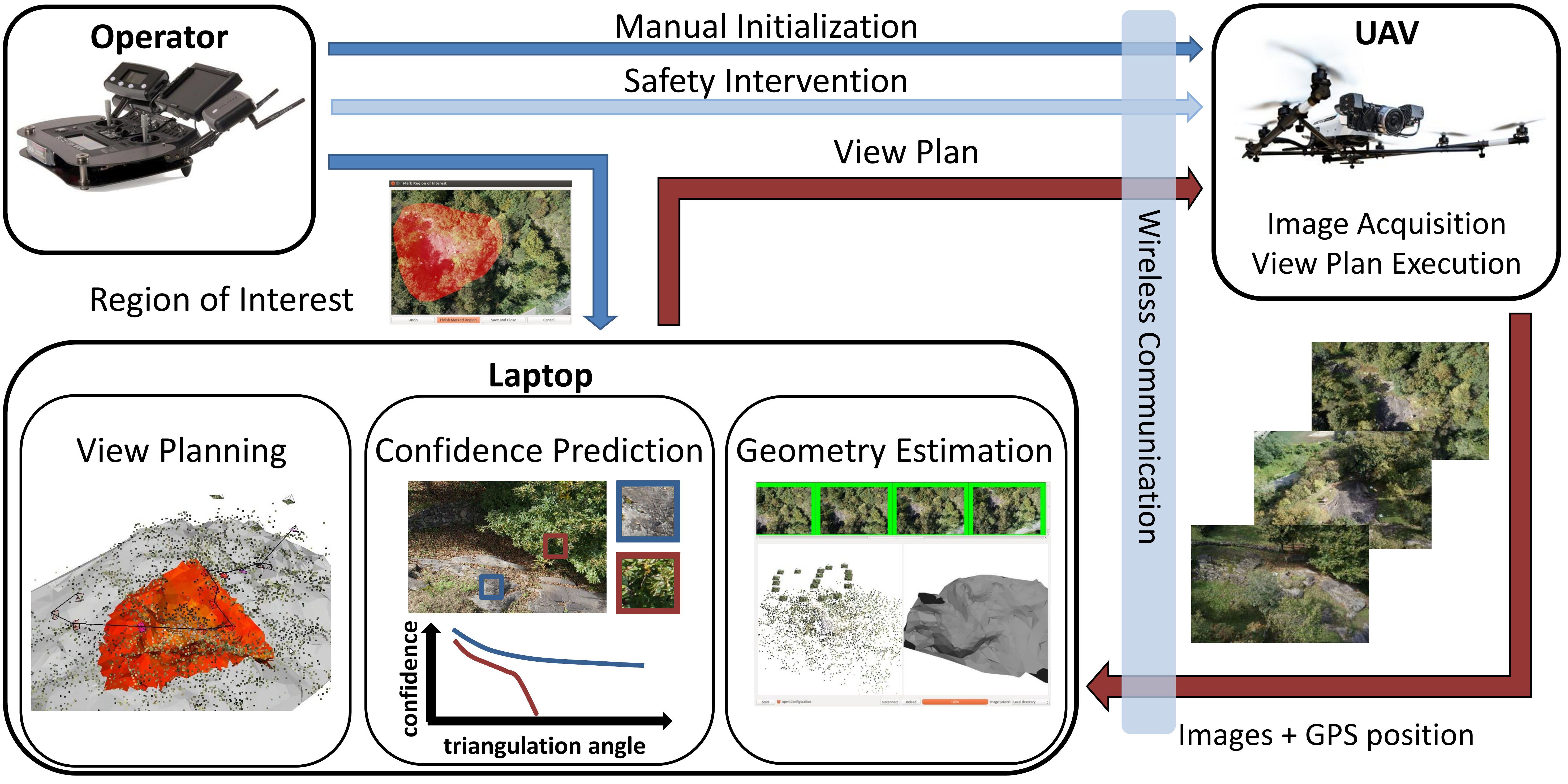} 
    
  \caption{Autonomous Image Acquisition. 
 After a manual initialization, our system loops between view planning and autonomous execution.
   Within the view planning procedure, we leverage machine learning to predict the best camera constellation 
 for the presented scene and a specific dense MVS algorithm.
 This MVS algorithm will use the recorded high resolution images
 to produce a highly accurate and complete 3D reconstruction off-site in the lab.
  }
   \vspace{-15pt}
  \label{fig:system}
\end{figure}

This MVS confidence prediction is related but not equal to the (two-view) stereo confidence prediction,
which is a topic of increasing interest in the domain of stereo vision~\cite{haeusler13,spyro14,park15,mostegel16}.
In stereo vision, the confidence encodes the likelihood that an already computed depth value is correct,
whereas in our case it encodes the likelihood that we will be able to compute a correct 3D measurement later-on.
Despite this difference, the training of both tasks is closely related 
and requires a large amount of data.
Up to now, obtaining this training data was a tedious and time consuming task,
evolving manual interaction~\cite{ladicky12,geiger12,menze15}, synthetic data~\cite{butler12,peris12,menze15} 
and/or 3D ground truth acquisition with active depth sensors~\cite{strecha08dataset,geiger12,menze15,scharstein14}.
In \cite{mostegel16}, we present a new way of obtaining this training data for stereo vision.
The main idea is to use multiple depthmaps (computed with the same algorithm) from
different view points and evaluate consistencies and contradictions between them to collect training data.
In this work, we extend this completely automatic approach to multi-view stereo.

After training, our system operates completely on-site (Fig.~\ref{fig:system}).
For estimating the scene geometry, we use the already acquired images for performing
 incremental structure-from-motion (SfM)~\cite{hoppe12onlinesfm} 
and incremental updates of an evolving mesh~\cite{hoppe13incmeshing}.
 Both modules run concurrently in real-time and deliver the camera poses of the acquired images
 and a closed surface mesh representation of the scene.

Based on this information,
we plan future camera positions that maximize the quality of 
a later-on dense 3D reconstruction.
This task falls in the domain of view planning, which has been shown to be NP-hard~\cite{tarbox95complete_coverage}.
Consequently,
a wide range of very task specific problem simplifications 
and solutions were developed in the communities of
robotics~\cite{sadat15,rainville15,nieuwenhuisen16,heng15,galceran13,galceran15,forster14,dornhege13,bircher15AR,alexis15,mostegel14icra,vasquez_gomez13raytracing,scott03view_planning_survey,
hollinger12vp_underwater,englot11inspection,schmid12view_planning_mav},
photogrammetry~\cite{martin16,liu14,hossein14a,hossein14b} and computer vision~\cite{pistellato15, scott02pose_error,trummer10nbv,hoppe12nbv,dunn09nbv,munkelt10mvp,haner11vslam_viewplanning}.
What kind of simplification is chosen strongly depends on the used sensor,
the application scenario and the time constraints.
In this work,
we propose a set of simplifications that allows us 
to compute a view plan in a fixed time-frame.
In contrast to active depth sensors, 
  a single 3D measurement in monocular 3D reconstruction requires multiple images to observe the same physical scene part.
Thus our first simplification is to 
remove this inter-camera-dependency by planning triplets of cameras as independent measurement units.
Second, we introduce the concept of surrogate cameras (cameras without orientation)
to reduce the dimensionality of the search space.
Finally, we lower the visibility estimation time through inverse scene rendering.
In contrast to the works above, our formulation allows us to evaluate a large number of potential camera poses
at low cost,
while the run-time can be adjusted to the acquisition requirements.

In the following, we first describe the training and setup of our MVS confidence predictor.
Then we describe our fixed-time view planning strategy.
In our experiments, we
evaluate the performance and stored information of the confidence predictor
on a challenging outdoor UAV dataset.
In the same domain, we finally evaluate our autonomous image acquisition system with respect to 
quality and completeness of the resulting 3D reconstructions.

\section{Multi-View Stereo Confidence Prediction}
\label{sec:confidence_prediction}
Given a specific scene structure (e.g. vegetation) and a camera constellation, 
the MVS confidence encodes
the likelihood that a dense reconstruction algorithm will work as intended.
With "work as intended" we mean 
that if a scene part is observed by a sufficient number of cameras then 
the algorithm should be able to produce a 3D measurement within the theoretical uncertainty bounds for each pixel that observes this scene part~\cite{mostegel16}.
The first matter we address in this section is how we can generate training data to predict the MVS confidence without any hard ground truth.
Therefore we extend our approach for stereo vision~\cite{mostegel16} to multi-view stereo.
Then we outline our machine learning setup and explain how we can use this setup to 
predict the MVS confidence in real-time during the image acquisition.

\subsection{MVS Training Data Generation}
As it is extremely tedious to come by 3D ground truth, the basic idea of \cite{mostegel16} is to
use self-consistency and self-contradiction from different
view points for generating labeled training data.
This approach is related to depthmap fusion, but outputs 2D label images instead of depthmaps.
Pixels that are associated with consistent depth values become positive training data,
while inconsistent depth values lead to negative training data.
This data is then used for training a pixel-wise binary classification task.
The main challenge during the training data generation is to keep 
the false positive rate (consistent but incorrect) and the false negative rate (correct but inconsistent)
as low as possible, while labeling as many pixels as possible.

In \cite{mostegel16}, we start by computing a depthmap for each stereo pair in the dataset.
A single depthmap can be interpreted as the 3D reconstruction of a camera cluster with two cameras and a fixed baseline.
In the case of multi-view stereo, we can choose an arbitrary number of cameras per cluster 
in any constellation.
As this general case has too many degrees of freedom to be estimated efficiently,
we limit ourselves to three cameras per cluster,
which is also the standard minimum number of cameras for most MVS approaches (e.g.~\cite{furukawa10pmvs,rothermel12}).
Within this triplet of cameras, the most important factor is the baseline between the cameras
or more precisely the triangulation angle between the cameras and the scene.
This triangulation angle can be freely chosen.
We want to use this property to learn the relationship between 
MVS confidence and the triangulation angle so that 
we can choose the right camera constellation for the presented scene in our view planning approach.
In theory, a large triangulation angle between cameras is beneficial as it reduces the 3D uncertainty.
However, in practice a large triangulation makes it more difficult to find correspondences between the images,
especially when the 3D structure is highly complex.
To learn this relationship, we first generate a large variety of triangulation angles in the training data.
We randomly sample image triplets from a fixed number ($t$) of triangulation angle bins,
while ensuring that the images have sufficient overlap.
For each of these camera triplets, we execute the chosen dense MVS algorithm 
and project the resulting 3D reconstruction back into the images to obtain one depthmap per image.
Using these depthmaps, we can proceed with the training data generation 
in three stages~\cite{mostegel16}.

The first stage has the purpose of reducing the influence of all consistent but incorrect measurements.
In practice, we can observe that the likelihood that two measurements of independent 3D reconstructions\footnote{3D reconstructions that were produced with the same MVS algorithm from independent image sets.}
are consistent but incorrect at the same time decreases as the relative view point difference increases.
Thus, we analyze how well each measurement is supported by reference reconstructions from a sufficiently different view point.
We treat a reference measurement as sufficiently different if the view angle difference $\alpha_\text{diff} > \alpha_\text{min}$
or the scale difference $s_{\text{res,query}} > s_{\text{min}}$ is sufficiently large.
We compute these values as 
$\alpha_{\text{diff}} = \measuredangle (\overrightarrow{\mathbf{p}_{\text{query}} \mathbf{c}_{\text{ref}} },\overrightarrow{\mathbf{p}_{\text{query}} \mathbf{c}_{\text{query}} })$
 and 
$s_{\text{res,query}} = \text{res}_\text{ref} / \text{res}_\text{query}  $ with $\text{res}_\mathsf{x} = f_\mathsf{x} / \|\mathbf{c}_\mathsf{x} - \mathbf{p}_{\text{query}}\|$,
where $\mathbf{c}_{\text{x}}$ is the mean camera center and $f_{\text{x}}$  the mean focal length of a camera triplet.
If a reference
measurement fulfills one of the two conditions, we increment the \emph{support} of the query measurement by one.
Note that as in~\cite{mostegel16}, reference measurements from a similar view point are
only allowed to increment the \emph{support} once.

\begin{figure}
  \centering
\includegraphics[width=0.8\columnwidth]{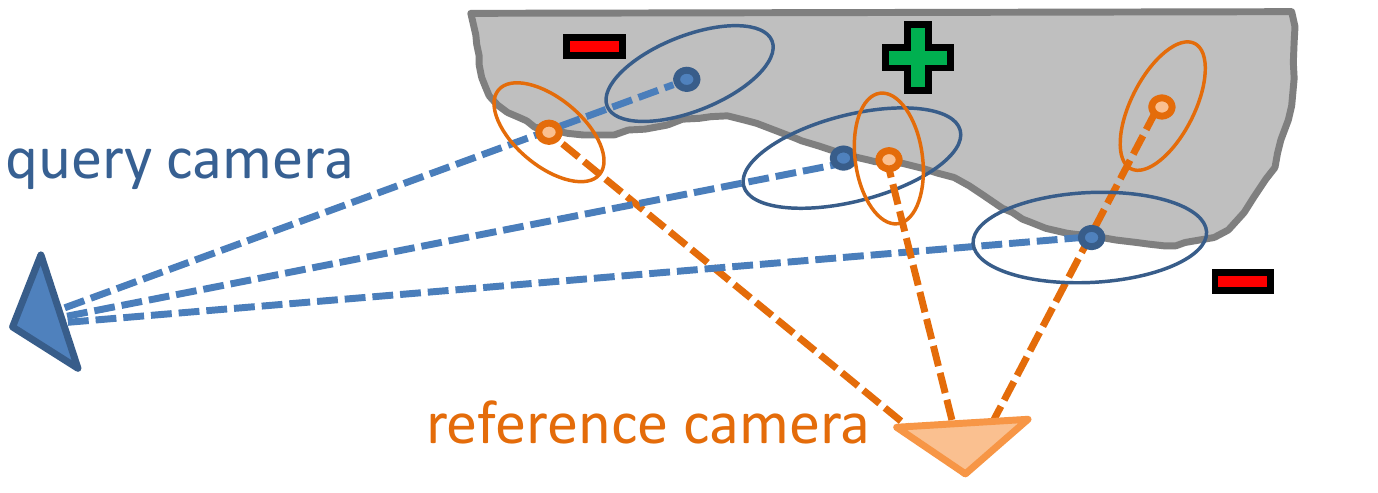} 
    
  \caption{Consistency voting. A positive vote (center) is  only cast
  if the reference measurement is within the uncertainty boundary of the query measurement.
  A negative vote is either cast if a reference measurement blocks the line of sight of the query camera (left) or 
  the other way around (right).}
  \vspace{-15pt}
  \label{fig:supp_n_contra}
\end{figure}

In the second stage, we let the parts of the depthmaps with at least one \emph{support} vote on the consistency of all depthmap values.
The voting process proceeds analog to~\cite{mostegel16}.
For each query measurement, we collect positive and negative votes as shown in Fig.~\ref{fig:supp_n_contra}.
The votes are weighted with their \emph{support} and their inverse 3D uncertainty~\cite{mostegel16}.
Based on the voting outcome, all pixels with at least one vote are then either assigned a positive or a negative label.

The third stage requires more changes to generalize to multi-view stereo.
While in~\cite{mostegel16} this stage only has the purpose of detecting outliers,
in our case we also have to detect missing measurements.
More precisely, we have to detect if the MVS algorithm
failed to produce any output in a region where it should have been geometrically possible
and use this case as a negative training sample.
For detecting these missing parts we use a combination of a depthmap augmentation~\cite{mostegel16} and 
two surface meshes.
We use two meshes with slightly different object boundaries to account for errors in the meshes.
To construct these meshes, we first use all available images in the dataset
to compute a sparse point cloud~\cite{rumpler2014automated}.
From this point cloud we robustly extract a surface mesh~\cite{labatut07delaunay,vu12dense_mvs},
and then shrink and expand this mesh for our purpose.
The shrunken mesh is obtained by 
 performing three iterations of neighbor-based smoothing. 
 In each iteration
 a vertex moves half the distance to the average position of the vertices that share an edge with this vertex.
For the second mesh, we expand the shrunken mesh again.
For this purpose, we compute a vector by averaging the motion vectors of a vertex 
and its neighbors from the shrinking procedure.
Each vertex is then moved twice the vector length in the 
opposite direction of this vector.
If the depthmap augmentations and the two meshes agree that some part of the scene is missing,
the corresponding pixels are used as negative training samples.

\subsection{Machine Learning Setup}
For view planning, we want to know which camera constellation will give us a good chance of 
getting a complete and accurate 3D reconstruction.
To help with this task,
we want to use the already acquired images during the acquisition.
For training, we pose the problem as a pixel-wise classification task.
During run-time, we compute the MVS confidence depending on the triangulation angle
and the scene around the pixel of interest.
For this task, we chose Semantic Texton Forests (STFs)~\cite{shotton08textonforest}.
We selected this approach for three main reasons.
First, this approach is very fast in the execution phase as it operates directly
on the input image (without any feature extractions or filtering).
Second, STFs have shown a reasonable performance in semantic image segmentation.
Third, it is possible to store meta information in the leaves of the forest.
We use this property to store the triangulation angle under which a sample was obtain (or failed to obtain).
This does not influence the learning procedure, but allows us to predict the reconstruction
confidence in dependence of the triangulation angle at evaluation time.

During the image acquisition, we want to compute the MVS confidence in real-time
on a specific computer for a specific high-resolution camera.
Thus we provide two ways to reduce the prediction time to the operator's needs.
First, we restructure the STF leaf nodes to contain a fixed number ($b$) of angular
bins with one confidence value for each bin.
Second, we can make use of the property that the confidence prediction is in general 
a smooth function for a specific type of object (see~Sec.~\ref{sss:valcamonica}).
Thus, we evaluate the MVS confidence on a regular grid and 
compute a confidence image with $b$ channels for each input image.

\section{View Planning}
\label{sec:planning}

\begin{figure}[top]
%   \vspace{10pt}
  \centering
\includegraphics[width=1\columnwidth]{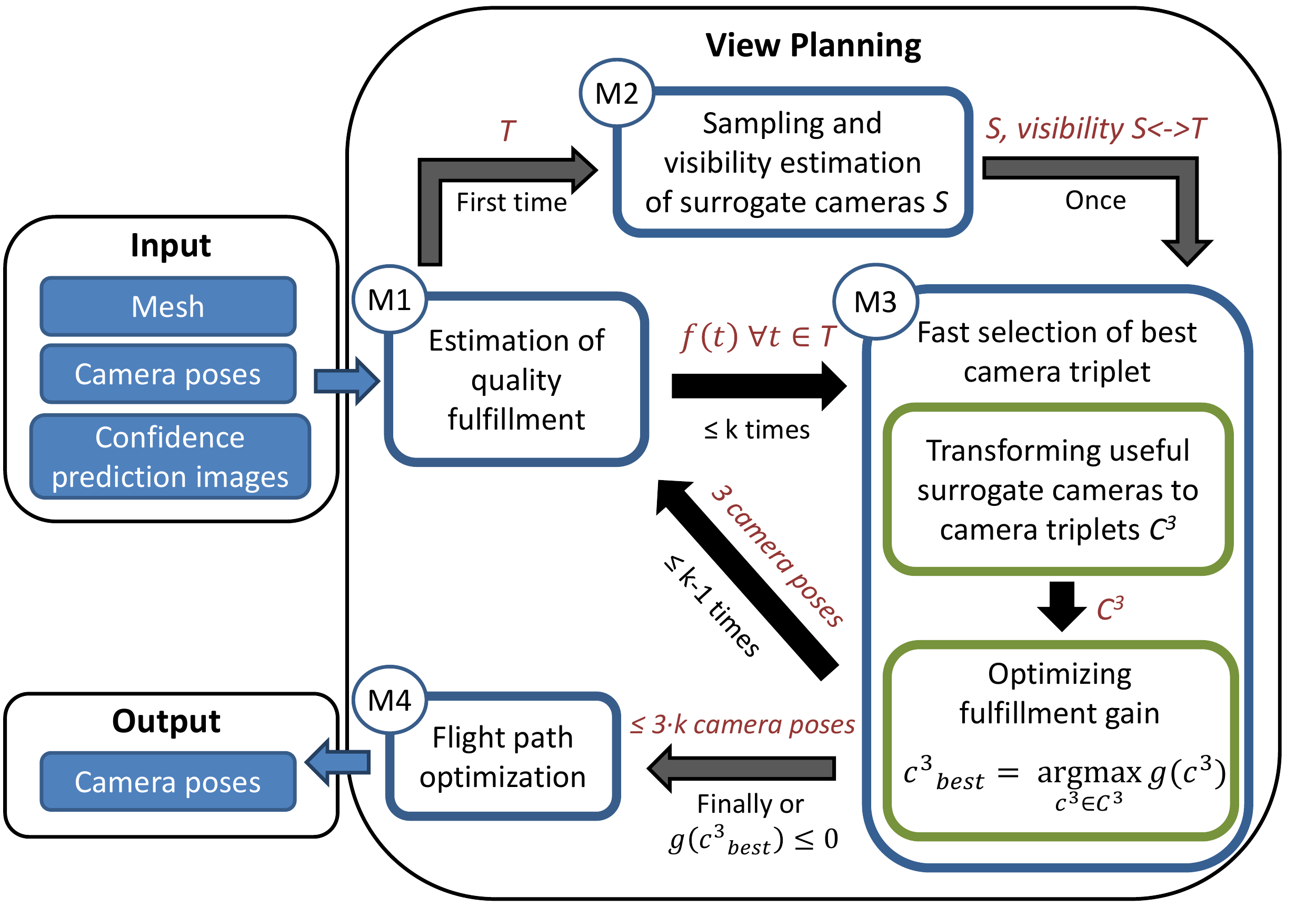} 
    
  \caption{View planning. Our algorithm tries to find the $k$ next best camera triplets for improving the
  acquisition quality. 
  Next to the arrows, we show the data communication between our submodules (M1-M4) in red and
  in black we show how often this data is computed.
  $S$ is the set of surrogate cameras, $T$ the set of considered unfulfilled triangles and $C^3$ the set of camera triplets generated from the surrogate cameras.
  }
   \vspace{-15pt}
  \label{fig:viewplanning}
\end{figure}

The aim of our view planning approach is to plan
a set of useful camera poses in a fixed time frame.
As the view planning problem is NP-hard, we have to make several 
simplifications to constrain the computation time.
One of our most prominent simplifications is that we plan equilateral camera triplets
instead of single cameras.
On the one hand, this lets us directly integrate our MVS confidence prediction and, 
 on the other hand, we can treat each camera triplet as an independent measurement unit.
In Fig.~\ref{fig:viewplanning} we show an overview of our approach,
which we use to guide the reader through our algorithm and its submodules.
As input our approach requires a snapshot of the estimated geometry (mesh and camera poses),
as well as the pre-computed MVS confidence images.
Further, the operator has to label a region of interest in one of the images (Fig.~\ref{fig:details}),
and define the desired quality constraints (ground resolution and 3D accuracy).

\textbf{Estimation of quality fulfillment (M1).} 
The aim of this submodule is to estimate how well our 
desired quality constraints are currently fulfilled by the already captured images.
For this estimation, we need the 
already acquired images and their camera poses $C_t$ as well as the surface mesh.
First, we bring all mesh triangles within the region of interest to approximately the same size 
through iteratively splitting them until the maximum edge length equals the 
average edge length before splitting.
Within the region of interest we then randomly select a fixed number $N_t$ of triangles.
Next we determine the visibility information between these triangles and $C_t$ through rendering the mesh.
Based on the information which cameras see which triangles, we evaluate
how well the desired quality constraints are currently fulfilled.
We compute the fulfillment separately for each triangle using 
four fulfillment functions.

(1) The \emph{coverage} is modeled as a
Boolean with $f_{cov} = 1$ if a triangle is visible in a minimum of $c$ cameras and $f_{cov} = 0$ otherwise.
(2) The \emph{resolution} requirement ($px/m^2$) is defined as
$f_{res} = \frac{r}{r_d}$ (truncated above 1) for a desired resolution $r_d$. 
(3) The fulfillment of the \emph{3D uncertainty} requirement is defined
as $f_{unc} = \frac{a_d}{\sqrt{u}}$ (truncated above 1)
 for a desired accuracy $a_d$. 
Here, $u$ stands for the maximum Eigen value of the covariance matrix related
to a triangle's centroid~\cite{hartley04multiview}.
(4) The last fulfillment function is the output of our MVS confidence prediction algorithm $f_{conf}$ (Sec.~\ref{sec:confidence_prediction}).

For evaluating these functions, 
we generate all possible combinations of camera triplets from the cameras that observe a triangle $t$ ($c^3 \in C_t^3$).
We then evaluate the combined fulfillment function as:
\begin{equation}
 f(t,c^3) = (\alpha f_{res} + (1-\alpha) f_{unc} ) \cdot f_{cov} \cdot f_{conf}
\end{equation}
This formulation allows the operator to define the relative weight $\alpha$ between 
desired ground resolution and 3D accuracy, while the coverage and MVS confidence
encode the chances of a successful reconstruction.
The overall fulfillment of a triangle $t$ is computed as $f(t) = \max_{c^3 \in C_t^3}{f(t,c^3)}\inlineeqno$. 

Based on the fulfillment information, we now further reduce the number of considered triangles
to a triangle set $T$.
We guide this reduction such that we end up with triangles that
have a low fulfillment but are well distributed over the scene of interest.
Thus, we randomly select a fixed number $N_v$ of triangles from a piece-wise constant distribution,
where the chance of selecting a triangle $t$ is weighted with $w(t) = 1 - f(t )/f_{conf} (t )$.
We remove $f_{conf}$ from the weight to
avoid bias towards structures that might not be reconstructible at all.

\textbf{Surrogate cameras (M2).} 
In this submodule, we use the concept of surrogate cameras
to estimate the visibility
of mesh triangles from a large number of possible camera positions.
Thus, we first 
randomly sample a fixed number $N_p$ of 3D positions in the free space of the scene.
These 3D positions represent the camera centers of surrogate cameras.
A surrogate camera has an unlimited field of view and thus also no orientation at this point 
(later we will transform this surrogate camera into an equilateral camera triplet).
The usage of surrogate cameras allows us to reformulate the visibility estimation problem
 and to estimate which surrogate cameras
are visible from a given triangle instead of the other way around.
The benefit of this formulation is that we are able to control execution time of the visibility estimation with the number of considered triangles
instead of the number of considered camera poses.
This enables us to evaluate a high number of camera positions at low cost.
For each triangle $t \in T$,
we place a virtual camera in the scene. The camera center of a virtual camera is set to the triangle's centroid and the
optical axis to the triangle's normal. We set the focal length of this camera such that we get a fixed field of view $\phi$. 
Now we use the virtual cameras for rendering the scene, i.e. the mesh and the 3D points that define the centers
of the surrogate cameras.
The resulting visibility links are stored in the surrogate cameras.

\textbf{Finding the best camera triplet (M3).} 
To find the best camera triplet at a low computational cost,
we guide the transformation from surrogate cameras to camera triplets
such that we only need to evaluate potentially useful and feasible camera constellations.
Thus, we first  compute the potential fulfillment gain $g_{pot}(t)$ of a surrogate camera
with respect to a linked triangle $t$.
Formally, we define $g_{pot}(t) = max_\alpha\{ f(t,c^3_\alpha) - f(t) , 0 \}$,
for a hypothetical equilateral camera triplet $c^3_\alpha$, that has the surrogate camera in its center
and where each camera directly faces towards the triangle.
The triangulation angle $\alpha$ defines the distance between the cameras in the $b$ steps of 
the predicted MVS confidence, which we evaluate with the confidence image 
of the closest already captured image (with respect to the surrogate camera) that observes the triangle.

Using this potential gain information, 
we determine in which direction the surrogate cameras should face.
Therefore, we perform a weighted mean
shift clustering on the rays towards the linked triangles. 
As a weight we use the fulfillment gain and the bandwidth is set to the minimum camera opening angle. The winning cluster (i.e. the
cluster with the highest potential fulfillment gain) is chosen to define the general viewing direction of the surrogate camera.
Then we update the visibility information and the potential gains of the 
now oriented surrogate cameras.

Given the orientation, we generate $b$ camera triplets for each 
surrogate camera,
one for each confidence bin.
For each camera triplet $c^3$ we efficiently check the distance to obstacles~\cite{lau13}
and compute the fulfillment gain of $c^3$ as
\begin{equation}
 g(c^3) = \sum_{t\in T_{c^3}} max\{ f(t,c^3) - f(t) , 0 \},
\end{equation}
where $T_{c^3}$ are the triangles that are visible from $c^3$.
Over all triplets,
we find the best camera triplet as
\begin{equation}
 c^3_{\text{best}} = \text{arg}\max_{c^3 \in C^3} g(c^3),
\end{equation}
where $C^3$ is the set of all generated camera triplets\footnote{
For the implementation, we can drastically reduce the number of evaluations
by using the potential gain.
If we start with the surrogate camera with the highest potential gain,
we can stop if $\sum_t g_{pot}(t)$ of the evaluated surrogate camera is zero or 
smaller than the current best gain.}.
If $g(c^3_{\text{best}})$ is greater than zero and we have not yet planned $k$ camera triplets, 
we add $c^3_{\text{best}}$ to the set of already acquired images ($C_t$) and plan a new camera triplet.
Otherwise, we pass all planned camera triplets with positive gain on to the flight path optimization.

\textbf{Flight path optimization (M4).}
This module minimizes the travel distance between the camera poses and
ensures that the resulting images can be registered by the geometry estimation module.
First, we reorder the camera poses with a greedy distance minimization using the last captured
image as a starting point.
Then we check if the taken images can be connected to the given set of images respecting the capture sequence.
We assume that this is the case if an image has a minimum overlap $o_{min}$ with at least one of the previously captured images.
If this is not the case we sample camera poses which fulfill this property along the 
trajectory from the closest previously captured camera pose to the target camera pose.
This results in a view plan that ensures a successful sequential registration of the planned image set.

\section{Experiments}
\label{sec:experiments}
We split our evaluation in two main parts.
The first part evaluates the performance and the information
which is stored by our confidence prediction approach.
The second part focuses on our autonomous acquisition system
and how it performs in a real world experiment.

\subsection{Confidence Prediction}

In the first part of this section we benchmark the performance of our training data generation and 
the prediction performance of the Semantic Texton Forest (STF)~\cite{shotton08textonforest} on the KITTI dataset~\cite{geiger12}.
In the second part, we use a challenging multi-view dataset to evaluate what the system can learn about two different
multi-view stereo approaches in relation to scene structure and camera constellation.

In all our experiments, we used the same STF setup.
We implemented the STF in the random forest framework
of Schulter~et~al.~\cite{schulter14a}.
We only use STF in its basic form (without image-level prior~\cite{shotton08textonforest}).
This means that the split decision is made directly on the 
image data (Lab color space) within a patch of the size $27\times 27$.
We trained 20 trees with a maximum depth of 20.
For the split evaluation, we used the Shannon Entropy, minimum leaf size for further splitting of 50,
5000 node tests, 100 thresholds and 1000 random training samples at each node.
For all our experiments, we extracted approximately 4 million training patches for each
class in training.

\subsubsection{KITTI2012 Dataset}
In this experiment, we apply our approach to the scenario of street-view dense stereo reconstruction
using the KITTI dataset~\cite{geiger12}, which provides a semi-dense depth ground truth recorded with a Lidar.

For learning, 
we follow the same procedure as in \cite{mostegel16} and
use the 195 sequences of 21 stereo pairs of the testing
dataset for automatically generating our label images.
We treat each stereo pair as a distinct cluster and 
use a semi-global matcher with left-right consistency check (SURE~\cite{rothermel12})
as the query algorithm.
As in \cite{mostegel16}, we evaluate the label accuracy and the average Area Under the Sparsification Curve (AUSC),
although with a slightly different setup.
While stereo confidence prediction \cite{mostegel16} tries to decide which depth values cannot be trusted from an already computed depthmap,
our aim is to predict which kind of structures cause more problems than others.
Thus, we remove all regions from the Lidar ground truth, which are not visible in both color images (including object occlusions).

With this setup we reach a labeling accuracy of \textbf{98.7\%} 
while labeling 35\% of the ground truth pixels (which is very similar to the results in \cite{mostegel16}).
For the sparsification we obtain a relative AUSC of 3.15 (obtained AUSC divided by optimal AUSC).
This means that the AUSC is \textbf{39\%} lower than random sparsification with 5.15.
This is a strong indication that the system learned to predict regions which are difficult to reconstruct
for the semi-global matcher.

For the matter of completeness, we also analyze the sparsification performance
of the STF~\cite{shotton08textonforest} with the exact same setup as in \cite{mostegel16}
(including the training data generation).
With this setup STF reaches a relative AUSC of 6.63.
It is not surprising that STF cannot reach the sparsification performance of stereo specific sparsification approaches (e.g. left-right difference with 2.81),
 as the STF only uses color information of a single image and thus has no chance to reason about occlusions.
Nevertheless, the STF was able to extract some high level knowledge in which regions the chances of failure are higher and 
thus still obtains a 31.4\% lower AUSC value than random sparsification (9.65).

\subsubsection{Val Camonica Dataset}
\label{sss:valcamonica}

\begin{figure*}[top]
     \vspace{-15pt}
  \centering
%\begin{minipage}{1\linewidth}
    \subfigure%[Reconstruction probability.]
    {

                \includegraphics[width=0.35\textwidth]{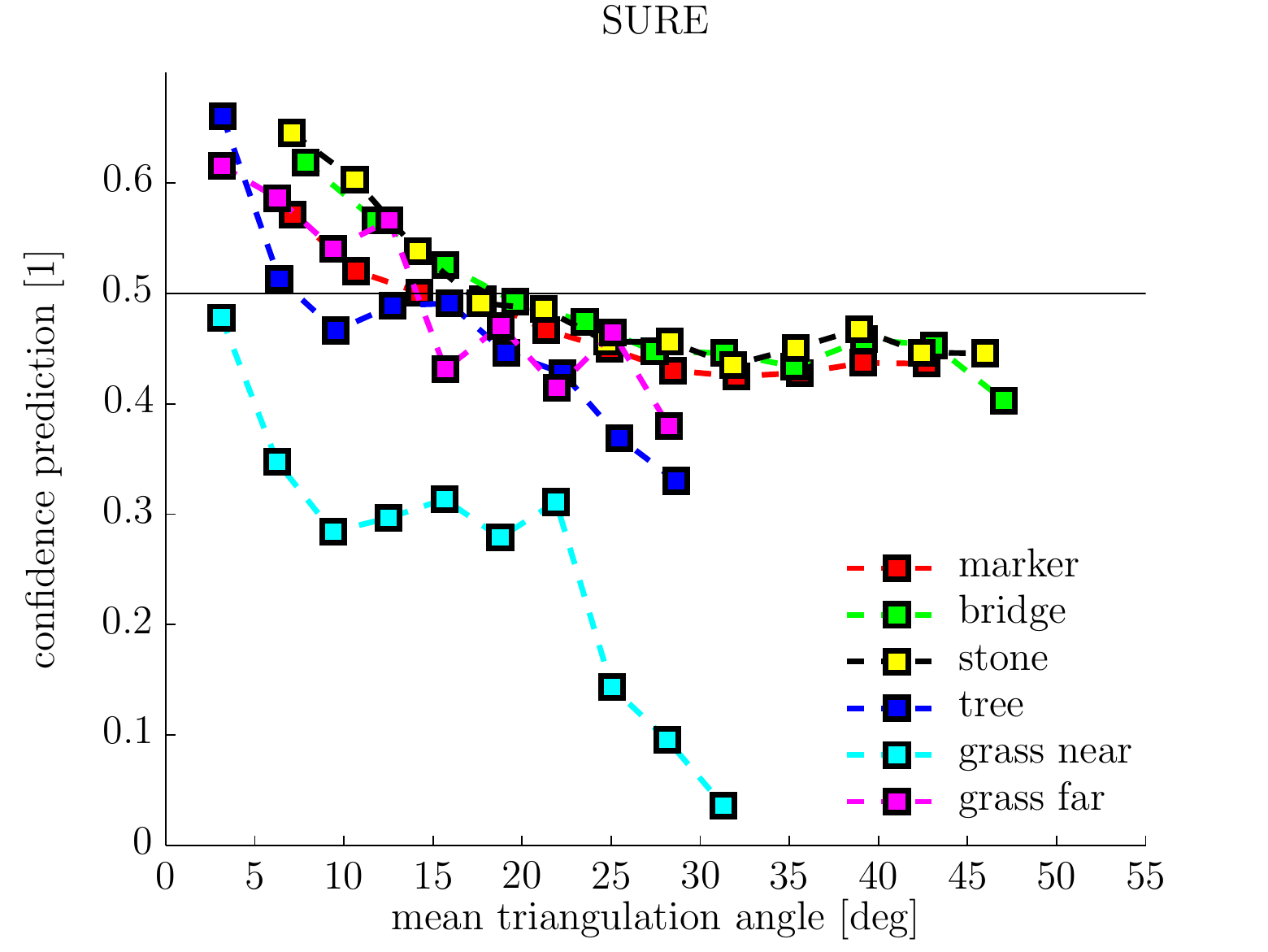} 
}\quad
    \subfigure%[Distribution.]
    {

                \includegraphics[width=0.35\textwidth]{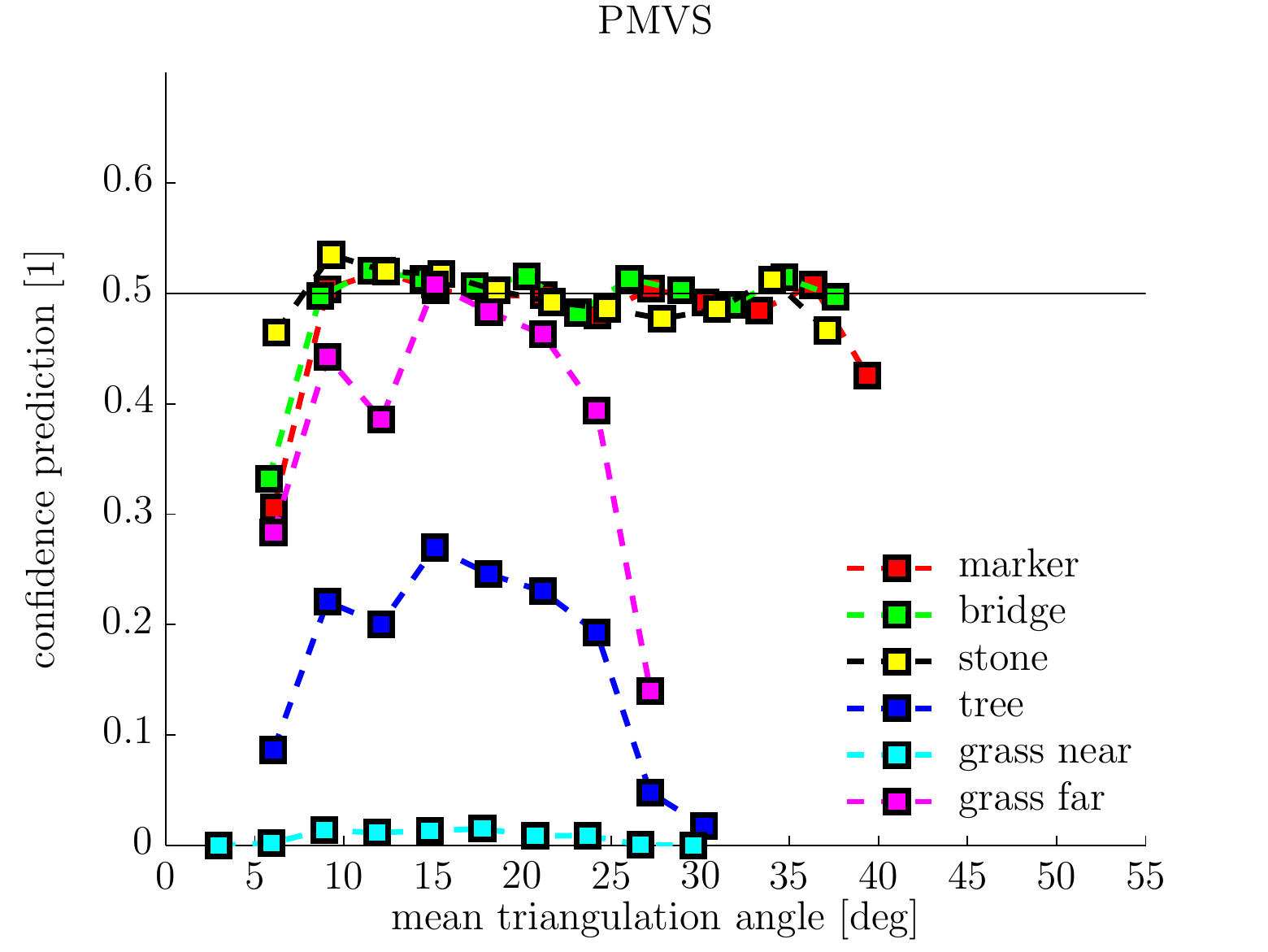} 
}\quad
    \subfigure%[Patches.]
    {
                 \includegraphics[width=0.12\textwidth]{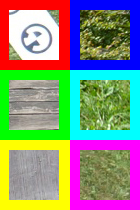}
}
%\end{minipage}
    \caption{Dependence of the confidence prediction on the triangulation angle and the 3D structure.
    On the right, we display the patches ($50\times50$px) which we used to produce the curves.
    These curves show the confidence prediction within angular bins (20 bins between min and max). 
    The curves stop if less than 1\% of the collected triangulation angles fall within a bin. 
    For both approaches (SURE and PMVS), there is a significant difference between smooth surfaces (marker, bridge, stone) and
    high frequency structures (tree, grass).
    The predicted confidence is to some extent correlated with the degree of non-planarity of 
    a structure.
    While grass viewed from far away is quite easy to reconstruct,
    the same grass viewed close up becomes very hard to reconstruct.
    For both approaches, the chance for reconstructing highly non-planar structures above $30^\circ$ is virtually zero.
    }
  \vspace{-10pt}
  \label{fig:sure_structures}
\end{figure*}

For the second dataset, we have chosen a reconstruction scenario in a 
closed real-world domain, where the task is the 3D reconstruction of prehistoric rock art sites
 in the Italian valley of Val Camonica.
The recorded dataset consists of over 5000 images of 8 different sites (see supplement), which 
contain a well-defined set of 3D structures (mainly rock, grass, trees, bridges, signs and markers).
These structures dominate nearly all sites in the region (hundreds), which makes
this a perfect example for learning and predicting domain specific properties of a query algorithm.

For generating camera triplets we used $t=5$ triangulation bins.
The lowest triangulation angle bin starts at a minimum angle of $4^\circ$ and
 ranges to double that value, where the next bin starts.
On each resulting triplet we execute a query algorithm three times at different image resolutions (levels 1, 2 and 3 of an image pyramid).
We evaluate two query algorithms for the dense 3D reconstruction.
The first query algorithm is based on semi-global matching SURE~\cite{rothermel12},
but can use more than two views for improving the reconstruction accuracy. 
In contrast, our second query algorithm PMVS~\cite{furukawa10pmvs}
 tries to densify an initial sparse 3D reconstruction
through iterative expansion.

For the quantitative evaluation of this experiment, we performed leave-one-out cross validation
across the 8 sites, i.e. we train on 7 sites and test on the remaining.
This led to the following classification accuracies:
PMVS: 81.1\% (STD: 4.2\%) and SURE: 65.3\% (STD: 6.1\%).
Within this context, we also analyzed the influence of regular grid sampling on the prediction performance.
For small grid sizes the classification error stays nearly the same (relative error increase is below 1\% for 4 pixels),
while for larger grid sizes it declines gradually (below 3\% for 16 pixels and below 7\% for 64 pixels). 
This means that regular sampling can drastically reduce the computational load of the prediction with only a small decrease of the prediction performance.

Now let us analyze what the system learned about the two algorithms
in relation to scene structures and triangulation angle.
In Fig.~\ref{fig:sure_structures} we show the confidence prediction for six different structures.
From this experiment we can draw several conclusions.
First, the 3D structure of the scene has a significant influence on how well
something can be reconstructed under a given triangulation angle.
The more non-planar a structure is, the harder it is to reconstruct at large triangulation angles.
Second, the two analyzed approaches react very differently to a change in triangulation angle.
While for SURE the confidence is always highest for very small angles,
PMVS' confidence stays constant
for smooth surfaces.
In the case of non-planarity, SURE is clearly more robust than PMVS.

\subsection{Autonomous Image Acquisition}
\begin{figure}
\centering
 \includegraphics[width=1\columnwidth]{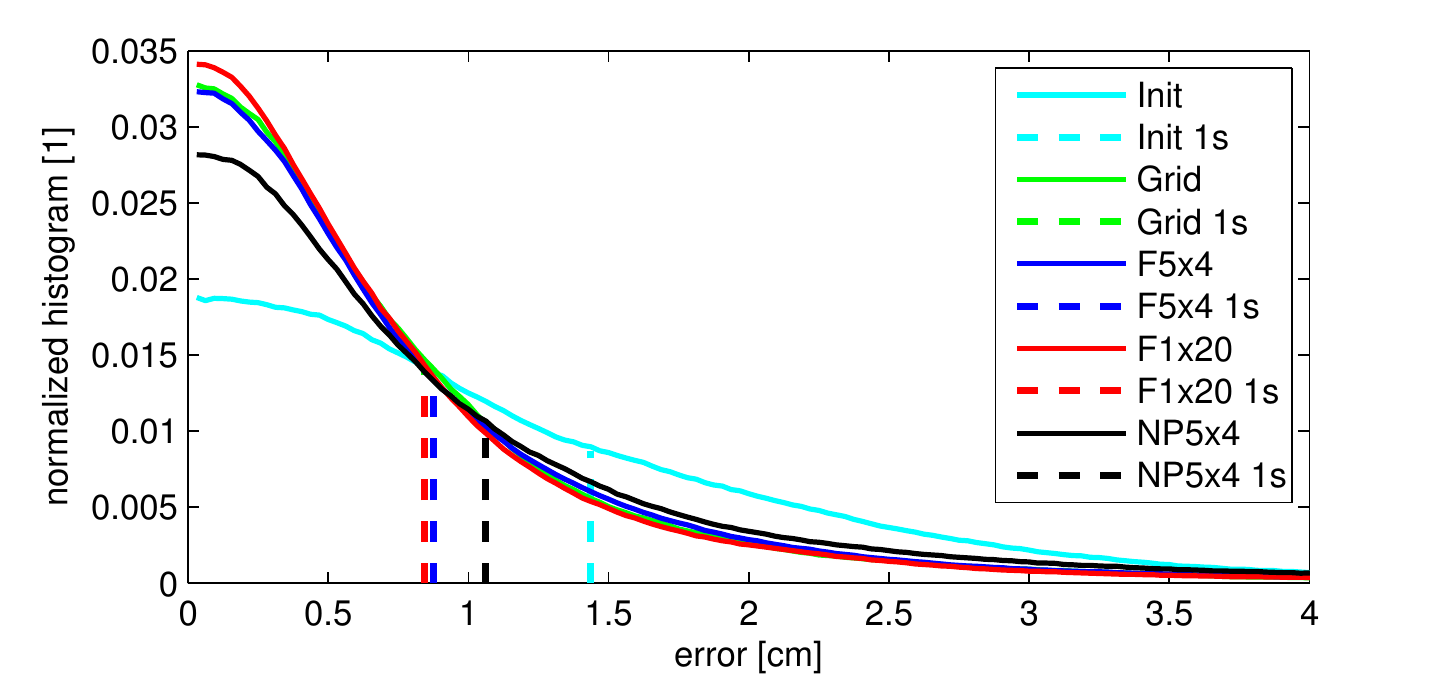} 
 \caption{Error histogram on the rock surface. We show the normalized histograms of the error distribution and the 
 1 $\sigma$ bound in which 68.3\% of all measurements lie. Grid and F1x20 share the same error bound.
 }
 \vspace{-15pt}
 \label{fig:error_histo}
\end{figure}

\begin{table*}
\centering
 \begin{tabular}[b]{|c||c|c|c|c|c||c|c|c|}
 \hline 
            &Init & Grid & F5x4 & F1x20 & NP5x4          & G+F5x4 & G+F1x20 & G+NP5x4\\\hline 
 \emph{cov} & $ 53.5 \pm 1.2 $ & $ 56.0 \pm 1.2 $ & $ \mathbf{65.6} \pm 1.6 $ & $ \mathbf{66.6} \pm 1.4 $ & $ 56.7 \pm 1.4 $ & $\mathbf{69.5} \pm 1.5$ & $67.0 \pm 1.5 $ & $57.2 \pm 1.2$\\\hline 
 $f_{res}$  & $ 17.9 \pm 1.3$ & $ 43.9 \pm 2.6 $ & $ 42.2 \pm 2.6 $ & $ \mathbf{47.5} \pm 2.7 $ & $ 29.3 \pm 2.3 $ & $\mathbf{52.8} \pm 2.7$ & $ \mathbf{55.3} \pm 2.7$ & $46.8 \pm 2.6$  \\\hline 
 $f_{unc}$  & $ 15.5 \pm 0.3 $ & $ \mathbf{22.8} \pm 0.4 $ & $ 21.2 \pm 0.5 $ & $ 20.7 \pm 0.5 $ & $ 19.9 \pm 0.5 $ & $\mathbf{27.7} \pm 0.5$ & $26.2 \pm 0.4$ & $25.6 \pm 0.4$ \\\hline 
 $f$        & $ 16.7 \pm 0.8 $ & $ \mathbf{33.4} \pm 1.5 $ & $ 31.7 \pm 1.5 $ & $ \mathbf{34.1} \pm 1.6 $ & $ 24.6 \pm 1.4 $ & $\mathbf{40.2} \pm 1.6$ & $\mathbf{40.7} \pm 1.6$ & $ 36.2 \pm 1.5$ \\\hline 
\end{tabular}
\caption{Fulfillment statistics in percent. We show the coverage of the region of interest \emph{cov}, the resolution fulfillment $f_{res}$ and the 
uncertainty fulfillment $f_{unc}$, as well as the overall fulfillment $f$ as defined in Sec.~\ref{sec:confidence_prediction}. 
We display the mean value and the standard deviation over the three surface meshes. We mark all results within the standard deviation of the best method
with a bold fond. In the first column we show the results with only the 19 initialization images, then we show the four standalone approaches.
The last three columns show a combination of the standard grid approach (Grid) with the other approaches.
}
\vspace{-10pt}
\label{tab:fulfillment}
\end{table*}

\begin{figure*}[t]
  \centering
\includegraphics[width=1\textwidth]{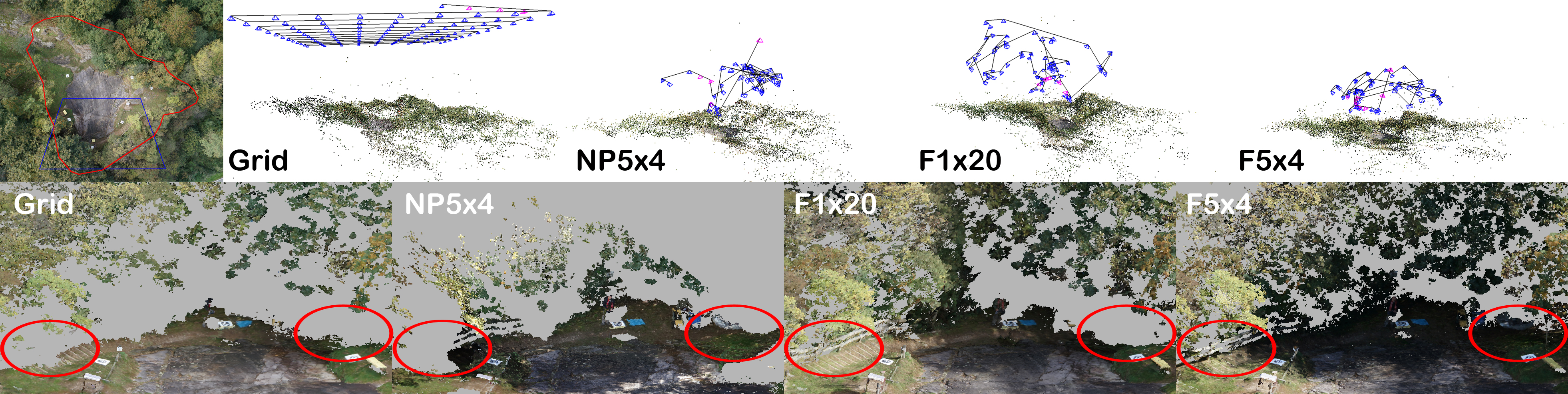} 
    
  \caption{Resulting 3D reconstructions. On the top left, we show one of the images acquired by the UAV with the region of interest in red.
  The other four columns show all view plans.
  The blue cameras are regular or triplet cameras, while the pink cameras ensure sufficient overlap for sequential registration.
  In the bottom row we show all four reconstructions (the field of view is marked blue in the top left image).
  Note that our approach has the best coverage underneath the trees. The point colors vary due to changing illumination conditions.
  }
  \vspace{-15pt}
  \label{fig:details}
\end{figure*}

To evaluate our image-acquisition approach in this scenario, we first run different view planning algorithms 
on-site and then analyze the effective reconstruction output, which is computed off-site.
As we also desire a reconstruction of the surrounding environment (which is dominated by vegetation),
we use SURE~\cite{rothermel12} as an MVS algorithm.
For this experiment, we run three versions of the proposed approach.
The first version is our full approach (F5x4), where we let the algorithm plan 4 camera triplets per iteration 
for a total of 5 iterations.
In the second version (F1x20), we let our approach plan the same number of total triplets (20) but in a single iteration, i.e. we disable the incremental geometry updates.
The third version (NP5x4) is exactly the same as F5x4 but without the prediction to constrain the triangulation angle.
As a baseline method, we use grid planning with 80 percent overlap.
All approaches share the same set of parameters.
The fulfillment requirements were set to $c = 3$, $r_d = 8$\,mm and $a_d = 8$\,mm with $\alpha = 0.5$.
The safety distance was set to 5\,m at a maximum octree~\cite{lau13} resolution of 2\,m
and the minimum camera overlap for registration to $o_{min}=50\%$.
The triangulation angle was binned in $b=9$ steps of $5^\circ$ from $0^\circ$ to $\gamma_{max} = 45^\circ$.
For the inverse visibility estimation we set the parameters such that the planning approximately takes 5 seconds per planned triplet, i.e. $N_t = 2000$, $N_p = 5000$ and $N_v=200$ with $\phi = 120^\circ$.
This parameters resulted in an effective execution time per triplet of 5.98 seconds (STD: 2.19) over all experiments on a HP EliteBook 8570w.
The confidence was evaluate on a regular grid with a step size of 8 pixels, which resulted in a confidence prediction time of $\sim$2\,sec/image.
We acquire the images with a Sony Nex-5 16Mpx camera mounted on an Asctec Falcon8 octocopter.

For this experiment, we focus on one site in Val Camonica, namely Seradina Rock 12C.
The rock surface (17$\times$13\,m) is covered with prehistoric rock carvings and
is partly occluded by the surrounding vegetation (Fig.~\ref{fig:details}).
We placed 7 fiducial markers in circle around the rock of interest and measured them with 
a Leica total station.
These markers can be automatically 
detected in the images and are used for geo-referencing the offline reconstructions~\cite{rumpler2014automated}.
Additionally, a ground truth mesh of the rock (not the surroundings) was obtained through terrestrial laser scanning (TLS) in the same coordinate system two years before.
The mesh has a resolution of 8\,mm edge length and the accuracy of the laser scanner (Riegl VZ-400) is 5\,mm.
We use this mesh to evaluate the resulting 3D uncertainty.

To evaluate the coverage and the requirement fulfillment, we first obtain a geo-referenced sparse reconstruction from all flights on the day of the experiment ($\sim$500 images).
Then we obtain three meshes, one based on~\cite{labatut07delaunay,vu12dense_mvs} and the two others as described in Sec.~\ref{sec:confidence_prediction}.
As we know that these meshes will contain errors, we only use these meshes as a guideline for the evaluation.
Within the region of interest, we split all triangles to have a maximum edge length of 8\,cm.
For each taken image, we first compute the triangle visibility.
Then we produce a depthmap from all SURE 3D points linked to the image.
If the measured depth is either larger than or within 24\,cm of the triangle depth, we accept the 3D point 
as a valid measurement of the triangle.
Based on the links of the 3D measurement, we then compute the fulfillment of the triangle analog to Sec.~\ref{sec:confidence_prediction}.
Finally, this results in a set of fulfillment and coverage scores over all triangles in the region of interest.

In field, all approaches were initialized with 19 images taken in grid at a height of 50\,m above the lowest point of the site.
The region of interest was marked in one of the initialization images, such
that it is centered on the rock and includes a few meters of the surrounding vegetation (Fig.~\ref{fig:details}).
Landing and take-off are performed manually, while the view plans are executed autonomously by the UAV.
\vspace{-10pt}
\paragraph{Results.}

For each of our approach variants, we executed SURE only on the three images of the triplets.
Like this we can evaluate the general success rate of view planning variants
in analyzing on which triplets SURE succeeded to produce any 3D output.
Without the confidence prediction the success rate is very low (\textbf{18\%} for \textbf{NP5x4}).
This shows the gap between theory and practice.
While in theory a large triangulation leads to a small 3D uncertainty,
the matching becomes much more difficult and only flat surfaces survive.
However, with the proposed confidence prediction we were able to reach a prefect
success rate for our full approach (\textbf{100\%} for \textbf{F5x4}),
and  still reached an acceptable success rate without the reconstruction updates (\textbf{80\%} for \textbf{F1x20}).

In Table~\ref{tab:fulfillment} we display the effective fulfillment statistics of all approaches in the region of interest.
Of the standalone approaches, F1x20 and Grid take the lead, but are closely followed by F5x4.
The worst performance was reached by NP5x4.
While the dense grid performs well on the overall fulfillment,
we can see a \textbf{10\% gap} in the scene coverage, where F5x4 and F1x20 lead with nearly equal results.
F1x20 performs slightly better than F5x4, because F1x20 found a sweet spot in the center above the rock for a single triplet
where it was able to drop below the tree line and acquire a close up of the rock.

If we combine the results of the dense grid (Grid) with the proposed approach, we achieve the overall best results.
All evaluated measures improve significantly, which is an indication of a symbiosis between the approaches.
This suggests that for the given scene (which is quite flat for many scene parts) an initial grid reconstruction
with a subsequent refinement with the proposed approach is recommended.
Note that if the scene complexity increases and a grid plan can no longer be executed safely (e.g. underneath a forest canopy or indoors),
our planning approach is still applicable.

If we take a look at the error distribution in relation to the ground truth of the rock surface (Fig.~\ref{fig:error_histo}),
we can see that our approach and grid planning achieve very similar results.
Note the Grid only covered 87.4\% of the rock surface, while all others covered significantly more:
F5x4 covered 97.9\%, F1x20 94.7\% and NP5x4 94.0\%. 
This is a very promising result, as we only allowed our approach to use the planned triplets and no combination between them,
while we put no such restrictions on the Grid approach.
Furthermore, many of the triplets focused on the surrounding vegetation and the overall number of acquired images by our approach is 
lower than for the Grid approach (60 vs. 108 images).
Thus, our approach achieved a high accuracy at a higher coverage 
with fewer images, which can also be observed visually in Fig.~\ref{fig:details} and the supplementary material.

\section{Conclusion}
In this paper we presented a novel autonomous system for acquiring close-range high-resolution images
that maximize the quality of a later-on 3D reconstruction.
We demonstrated that this quality strongly depends on the planarity of 
the scene structure (complex structures vs. smooth surfaces), the camera constellation and the chosen dense MVS algorithm.
We learn these properties from unordered image sets without any hard ground truth and use the acquired knowledge
to constrain the set of possible camera constellations in the planning phase.
In using these constraints, we can drastically improve the success of the image acquisition,
which finally results in a high-accuracy 3D reconstruction with a significantly higher scene coverage
compared to traditional acquisition techniques.

\FloatBarrier
{\small
\bibliographystyle{ieee}
%\bibliography{egbib}
\bibliography{\BibPath/icg_abbrevs,\BibPath/user-bibtex}
}

\clearpage
\appendix
\section{Supplementary Material}
In this document we provide further figures such that the reader
can have a broader perspective of the conducted experiments and their outcome.

\begin{figure*}[t]
  \centering
\includegraphics[width=1\textwidth]{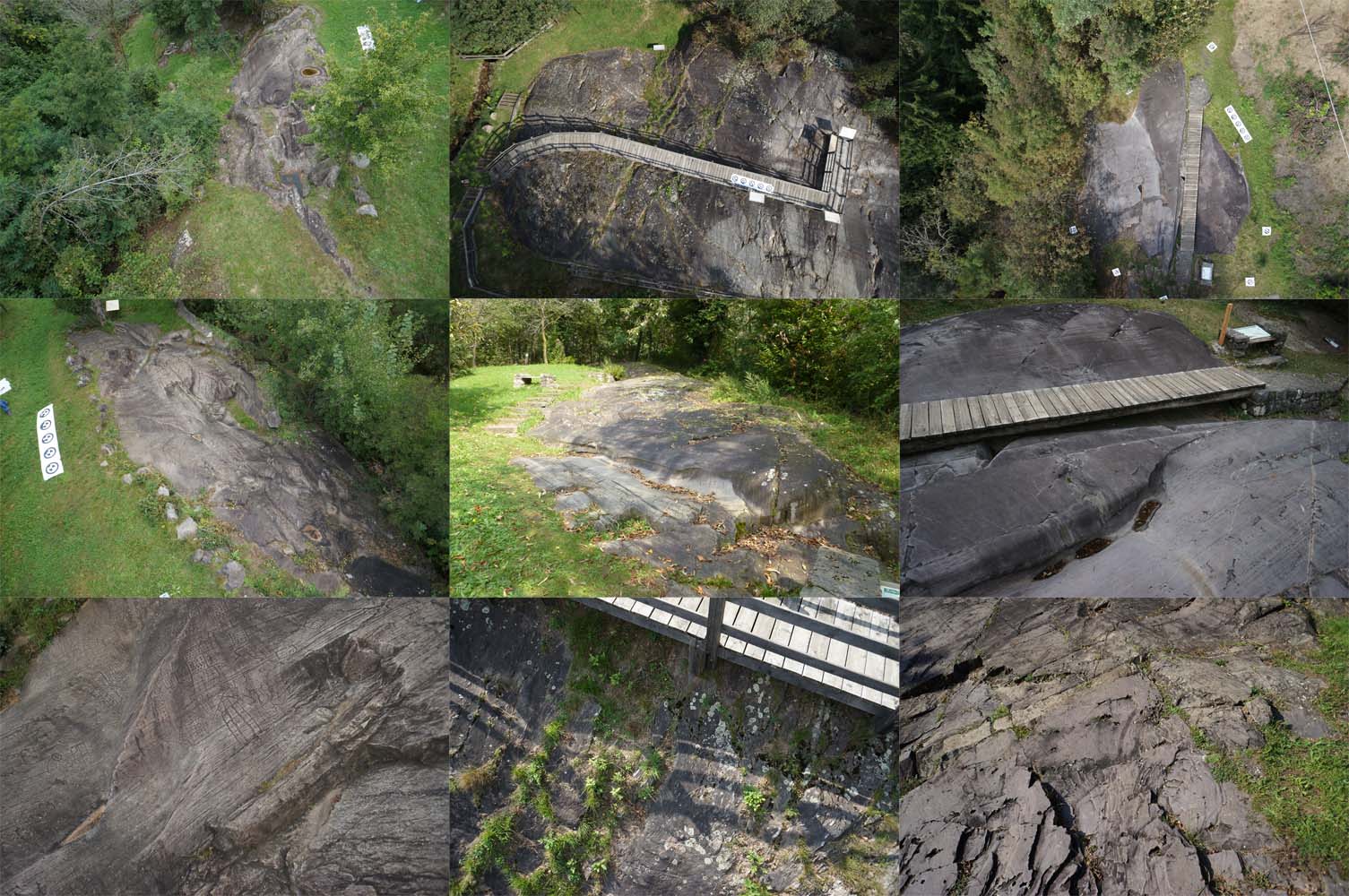} 
    
  \caption{Examples of the Val Camonica Dataset. In the top row we show three images from far away,
  in the middle three slanted views and in the bottom row three close-ups. All sites contain a limited set of 3D structures
  (mainly rock, grass, trees, bridges and markers).}
  \vspace{-15pt}
  \label{fig:sites}
\end{figure*}

In Figure~\ref{fig:sites} we show some examples of the Val Camonica Dataset which was used for training.
The whole dataset contains over 5000 images. 
The images contain a great variety of viewing angles and acquisition scales.
The camera to scene distance varies from to 2 to 50 meters.

\begin{figure*}[t]
  \centering
\includegraphics[width=0.6\textwidth]{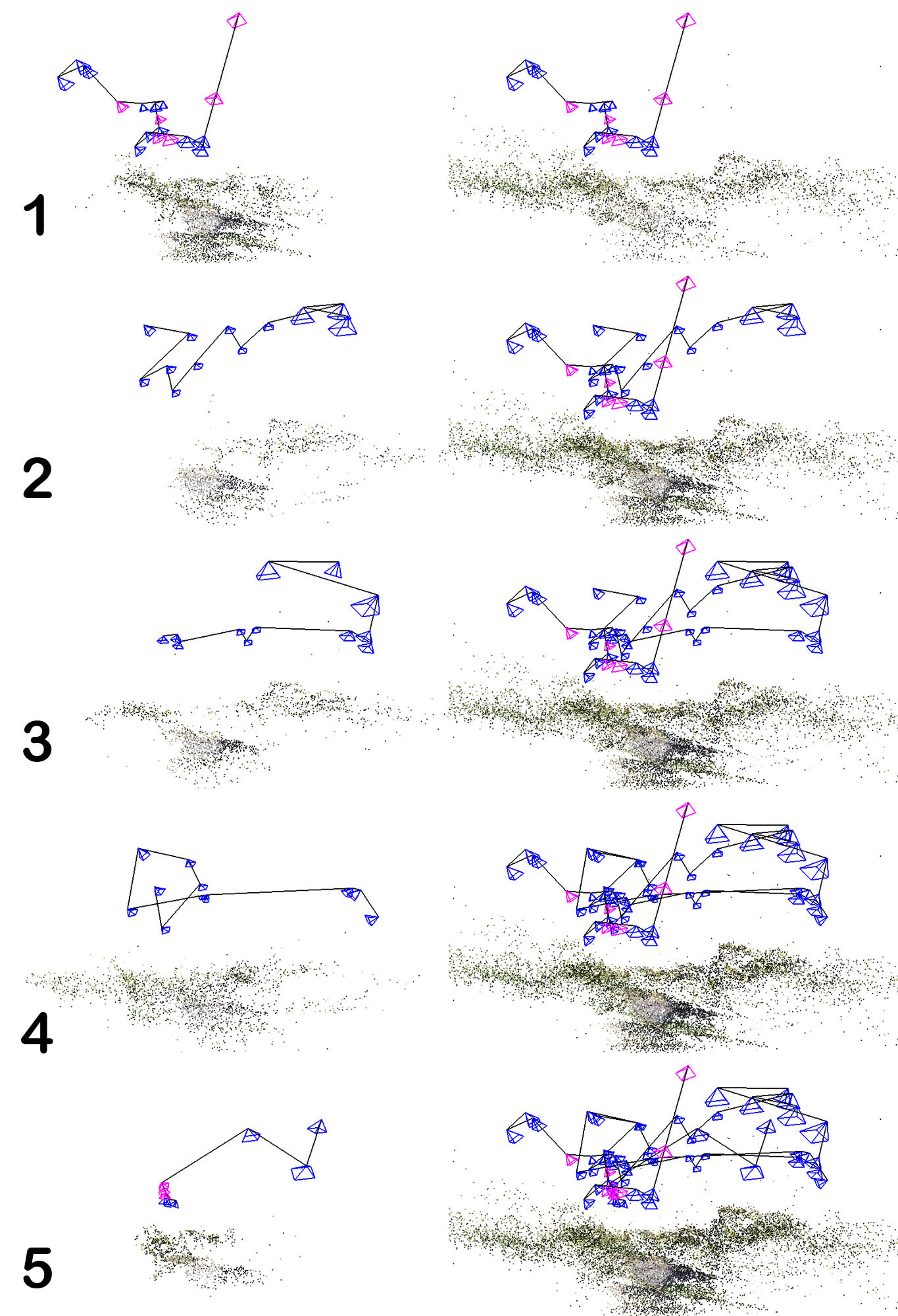} 
    
  \caption{View plans of our full approach (F5x4). From top to bottom we show the five planning iterations.
  On the left side, we show the current view plan together with the 3D points that were added to the sparse reconstruction with these images during the acquisition.
  On the right side, we show the sparse reconstruction at the time of planning and all computed view plans up to this point.
  The blue cameras belong to the planned camera triplets, whereas the pink cameras ensure a successful registration during the acquisition.}
  \vspace{-15pt}
  \label{fig:rf_4_evo}
\end{figure*}

\begin{figure*}[t]
  \centering
\includegraphics[width=0.5\textwidth]{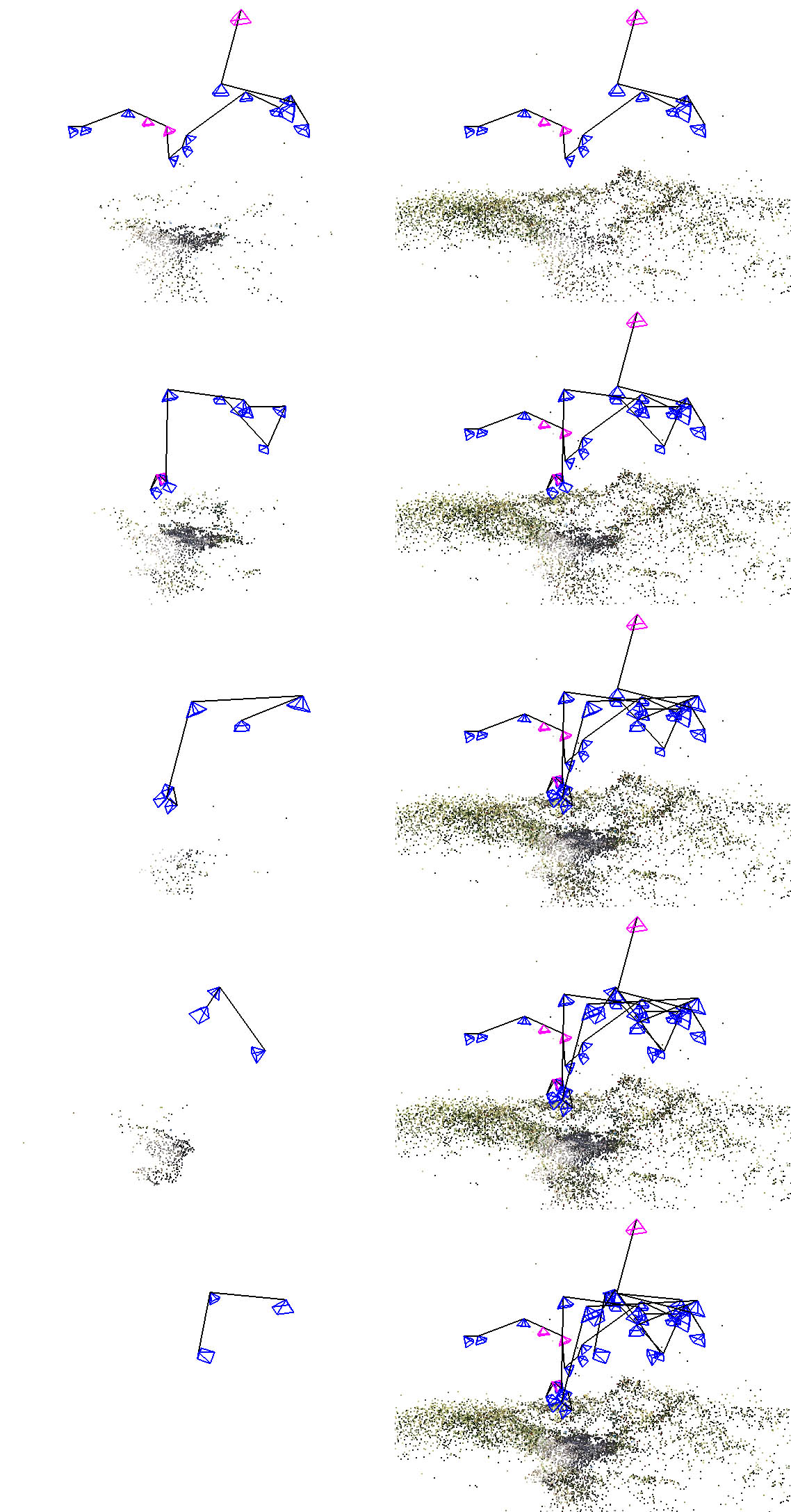} 
    
  \caption{View plans of the view planning approach without confidence prediction (NP5x4). From top to bottom we show the five planning iterations.
  On the left side, we show the current view plan together with the 3D points that were added to the sparse reconstruction with these images during the acquisition.
  On the right side, we show the sparse reconstruction at the time of planning and all computed view plans up to this point.
  The blue cameras belong to the planned camera triplets, whereas the pink cameras ensure a successful registration during the acquisition.
  Note that the algorithm constantly reduces the number of planned triplets as it has too much trust in the reconstruction algorithm.
  Further, the algorithm tries to push the triangulation angle to the allowed limits ($45^\circ$) as a large triangulation angle 
  decreases the theoretic 3D uncertainty.
  }
  \vspace{-15pt}
  \label{fig:no_rf_evo}
\end{figure*}

\begin{figure*}[t]
  \centering
\includegraphics[width=0.5\textwidth]{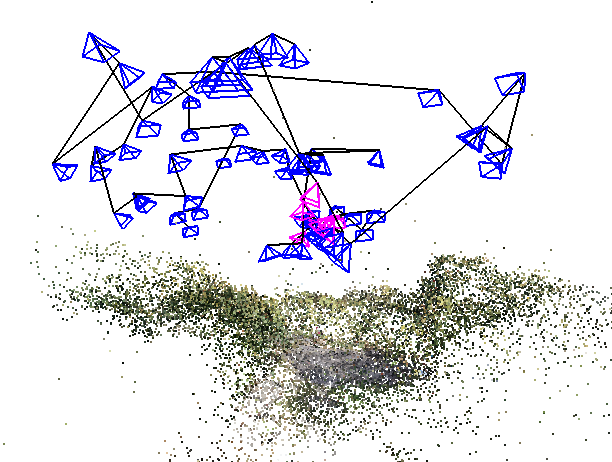} 
    
  \caption{View plans of our approach without structure updates (F1x20).
  In this version the approach only planned a single view plan.
  We show the view plan together with all 3D points that were generated on-site during the acquisition.
  The blue cameras belong to the planned camera triplets, whereas the pink cameras ensure a successful registration during the acquisition.}
  \vspace{-15pt}
  \label{fig:rf_20_evo}
\end{figure*}

\begin{figure*}[t]
  \vspace{-15pt}
  \centering
\includegraphics[width=0.98\textwidth]{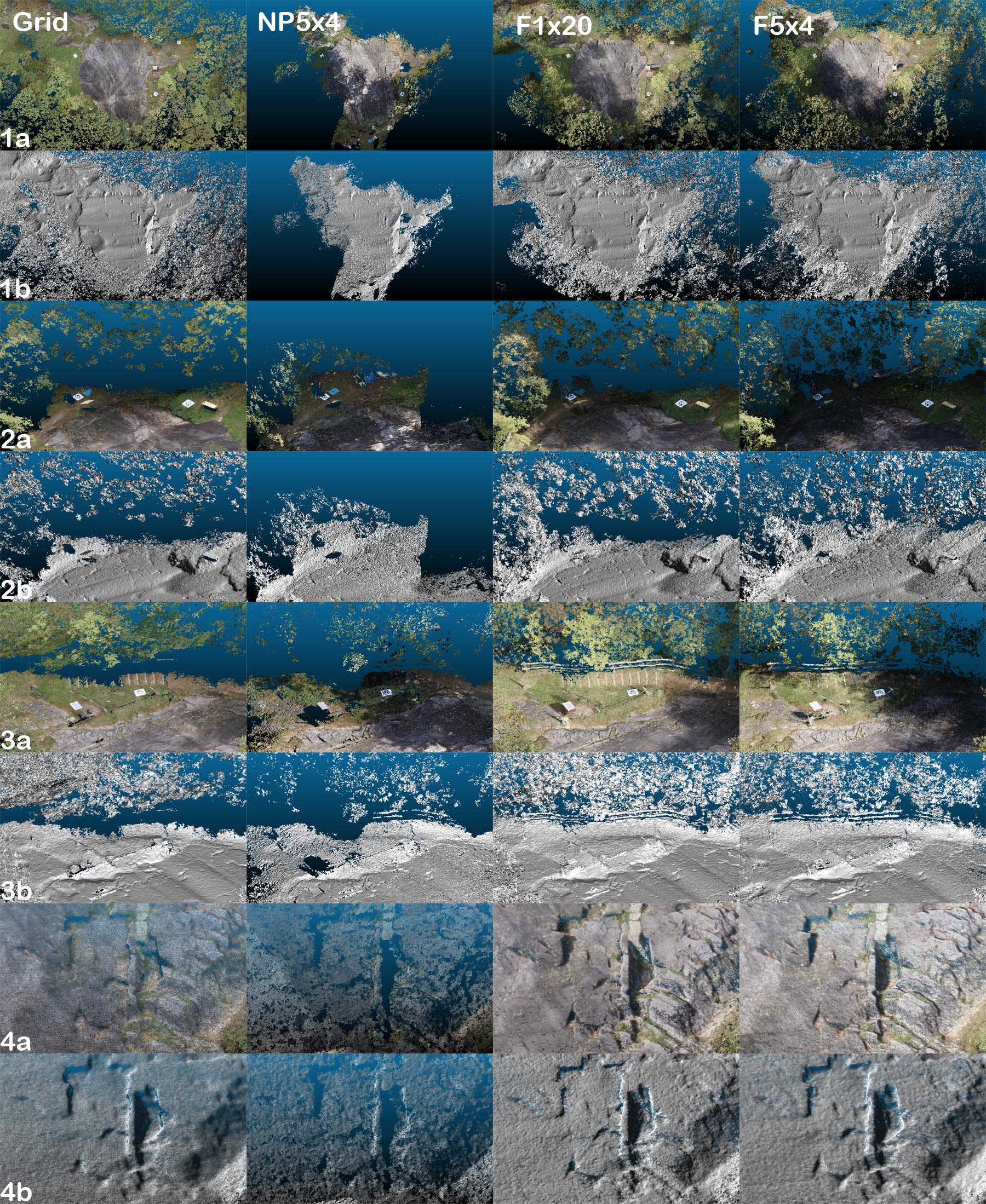} 
    
  \caption{We show the resulting reconstructions for the four different approaches.
  To eliminate the influence of color in the depth perception, we computed the normals of the point cloud and
  show the resulting point clouds also without color.
  For all triplet based approaches (NP5x4, F1x20, F5x4) we only show the output of the planned triplets (without initialization images or additional images for registration).
  Note that the coverage underneath the trees is significantly higher with confidence prediction (F1x20 and F5x4).
  The bottom row (4a-b) shows a close-up of the rock surface. Note that Grid is overly smooth, while F1x20 and F5x4 have much sharper edges.
  NP5x4 has a lot of missing parts caused by self-occlusion of the rock surface.
  }
  \vspace{-15pt}
  \label{fig:details4}
\end{figure*}

In Figures \ref{fig:rf_4_evo}, \ref{fig:no_rf_evo} and \ref{fig:rf_20_evo} we show the planning behavior of the 
three variants of the proposed approach together with the growing sparse reconstruction during the flights.
Note the different planning behavior between the variants.
With the MVS confidence prediction (F5x4 and F1x20) we see that the view planning algorithm
strongly varies the triangulation angle.
For smooth surfaces (the gray rock in the center) the algorithm prefers large angles as this decreases
the 3D uncertainty, but  it prefers a small triangulation angle for vegetation which
cannot be reconstructed under a large triangulation angle.
In contrast, if we execute our approach without the confidence prediction (NP5x4), we can see that 
the algorithm always uses a large triangulation angle, which leads to an incomplete 3D reconstruction where
only large smooth surfaces survive (see Fig.~\ref{fig:details4}).

\begin{figure*}[t]
  \centering
\includegraphics[width=1\textwidth]{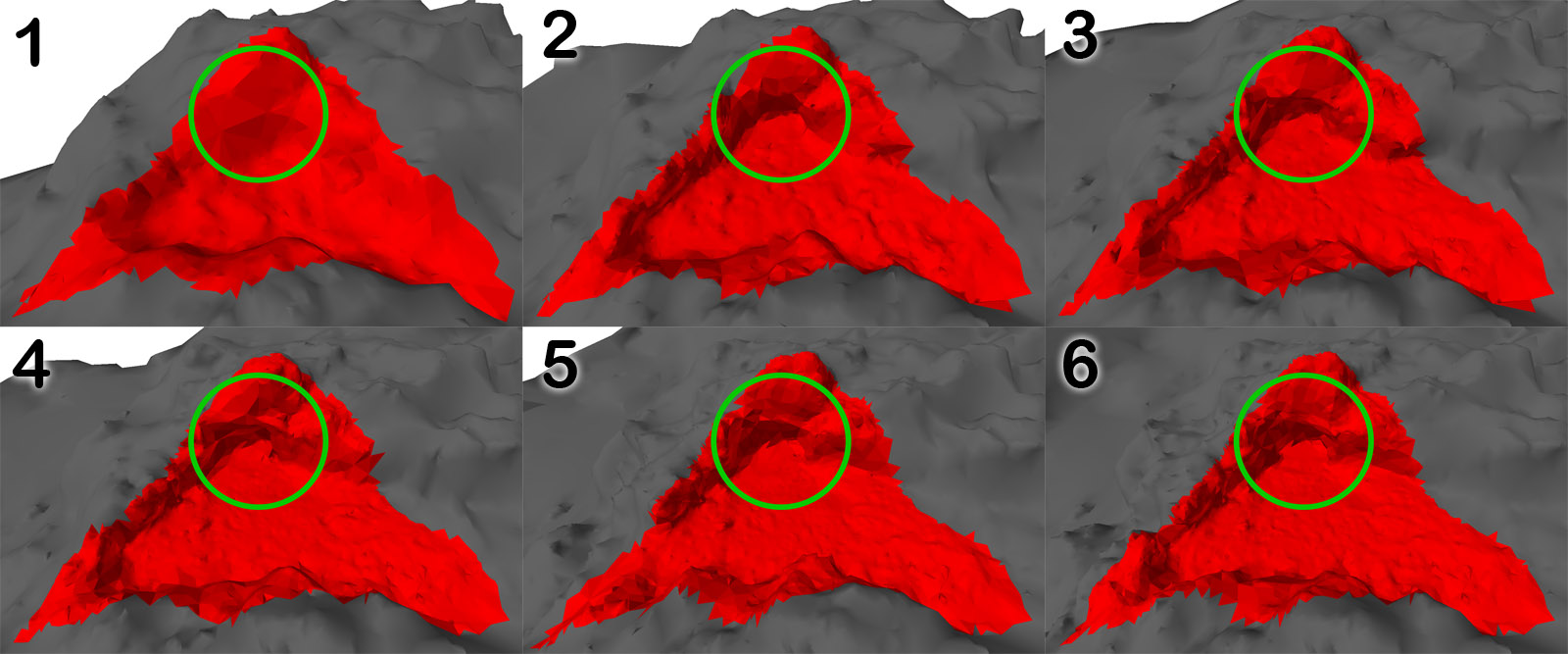} 
    
  \caption{Mesh evolution during the flight of our full approach F5x4. The mesh is colored red inside the region of interest and gray otherwise. 
  The green circle highlights a region where the mesh topology changes drastically during the acquisition.
  The biggest change can be observed in the first two iterations.
  The images 1 to 5 show the mesh at planning time of the corresponding iteration, whereas image 6 shows the mesh after the last plan execution.
  }
  \vspace{-15pt}
  \label{fig:mesh_evo}
\end{figure*}

In Figure~\ref{fig:mesh_evo} we show the evolution of the surface mesh over multiple iterations for our 
full approach (F5x4). We can see that the surface topology changes drastically for initially occluded 
parts. If the system would simply trust the initial reconstruction, the quality estimation in the 
marked region would be far from realistic.

In Figure~\ref{fig:details4} we show the resulting 3D reconstructions with and without color.
Note that the confidence prediction leads to a much higher scene coverage especially underneath the vegetation.

\begin{figure*}[t]
  \centering
\includegraphics[width=1\textwidth]{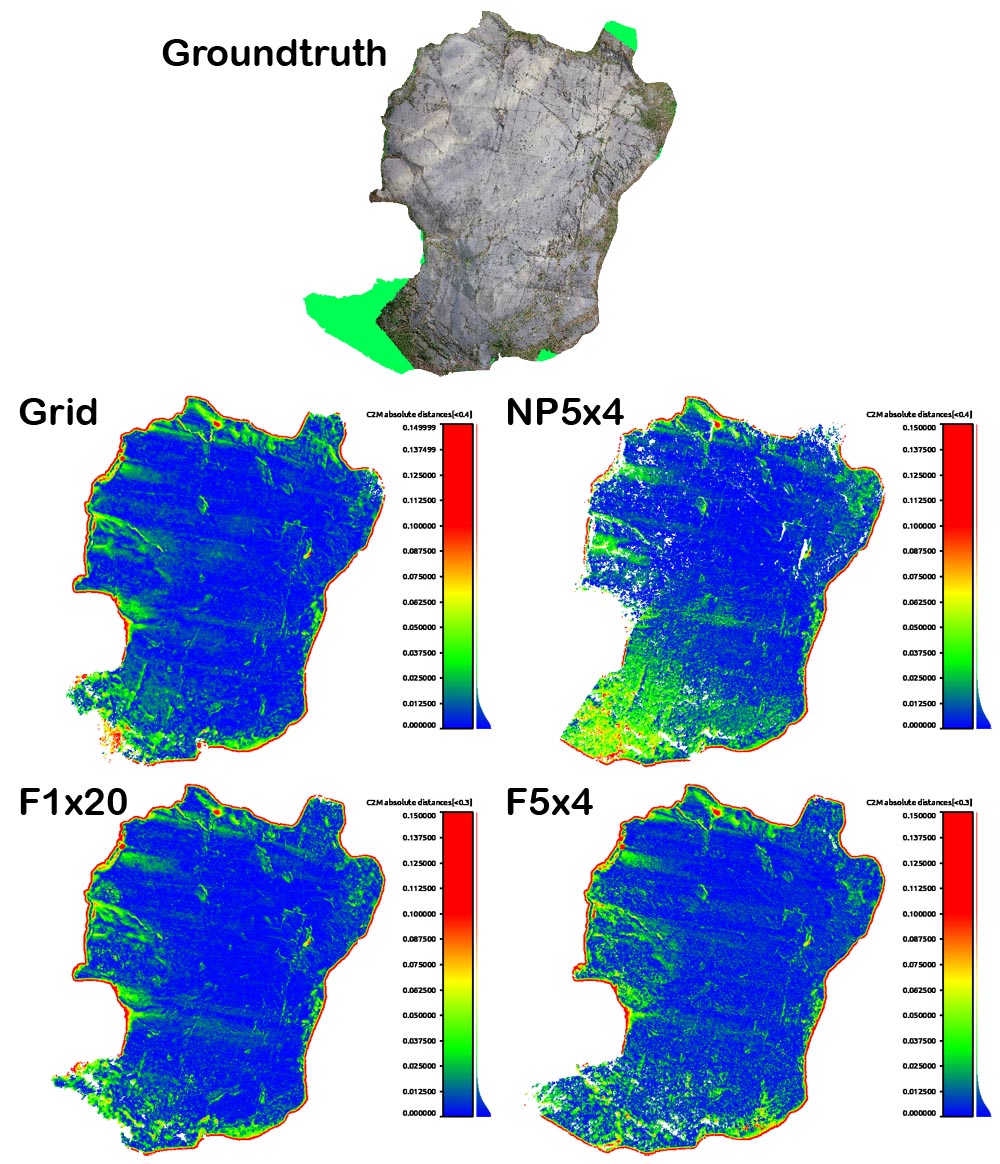} 
    
  \caption{Distance to the groundtruth. 
  On the top we show the groundtruth mesh which has been acquired with terrestrial laser scanning and was partly textured with UAV acquired images.
  The other images show the color coded distance of the final reconstructions to the groundtruth mesh.
  Note that the error of Grid, F1x20 and F5x4 is very similar, while NP5x4 has a much larger error in the bottom part of the rock.
  Further notice that F5x4 has the largest coverage of the rock surface in the lower left corner of the rock.}
  \vspace{-15pt}
  \label{fig:gt_comp}
\end{figure*}

In Figure~\ref{fig:gt_comp} we compare the resulting reconstructions to the groundtruth mesh of the rock surface.
We can see that all approaches except NP5x4 have a very similar error distribution.
Further, we can see that our full approach (F5x4) has the highest coverage of the rock surface, while
at same time sharing the high 3D reconstruction accuracy with Grid and F1x20.
\thispagestyle{empty}

\end{document}